\documentclass[dvipsnames,format=sigconf,natbib=false]{acmart}

\usepackage[separate-uncertainty=true, group-digits=integer]{siunitx}
\usepackage{booktabs,makecell,multirow}
\renewrobustcmd{\boldmath}{}
\newrobustcmd{\B}{\fontseries{b}\selectfont}

\usepackage[utf8]{inputenc}
\usepackage{dsfont}
\usepackage{amsmath}

\usepackage{yhmath}

\usepackage{algorithm}
\usepackage{algpseudocode}

\usepackage{comment}

\newtheorem{definition}{Definition}

\usepackage{tikz}
\usetikzlibrary{arrows,shapes,automata,backgrounds,petri,chains}
\usetikzlibrary{matrix,decorations.pathreplacing,calc,positioning,fit,trees}

\definecolor{graph1}{HTML}{05C090}
\definecolor{graph2}{HTML}{B30033} 

\usepackage[caption=false, font=footnotesize]{subfig}

\def\SMHD{\textsc{SMHD$_\mathbb{G}$}}
\def\FSim{\textsc{FSim$_\mathbb{G}$}}
\def\Ea{$\mathcal{E}_{\mathcal{G}^+}$}
\def\Eb{$\mathcal{E}_{\mathbb{G}}$}
\def\Ec{$\mathcal{E}_{\mathbb{G}}^*$}

\AtBeginDocument{%
  }

\copyrightyear{2025}
\acmYear{2025}
\setcopyright{acmlicensed}\acmConference[GECCO '25]{Genetic and Evolutionary Computation Conference}{July 14--18, 2025}{Malaga, Spain}
\acmBooktitle{Genetic and Evolutionary Computation Conference (GECCO '25), July 14--18, 2025, Malaga, Spain}
\acmDOI{10.1145/3712256.3726333}
\acmISBN{979-8-4007-1465-8/2025/07}

\RequirePackage[
  datamodel=acmdatamodel,
  style=acmnumeric,
  ]{biblatex}

\begin{document}

\title{Genetic Algorithms for Tractable Bayesian Network Fusion via Pre-Fusion Edge Pruning}

\author{Pablo Torrijos} 
\affiliation{%
  \institution{Universidad de Castilla-La Mancha}
  \city{Departamento de Sistemas Informáticos, Albacete}
  \country{Spain}}
  \email{Pablo.Torrijos@uclm.es}

\author{José A. Gámez} 
\affiliation{%
  \institution{Universidad de Castilla-La Mancha}
  \city{Departamento de Sistemas Informáticos, Albacete}
  \country{Spain}}
  \email{Jose.Gamez@uclm.es}

\author{José M. Puerta} 
\affiliation{%
  \institution{Universidad de Castilla-La Mancha}
  \city{Departamento de Sistemas Informáticos, Albacete}
  \country{Spain}}
  \email{Jose.Puerta@uclm.es}

\author{Juan A. Aledo} 
\affiliation{%
  \institution{Universidad de Castilla-La Mancha}
  \city{Departamento de Matemáticas, Albacete}
  \country{Spain}}
  \email{JuanAngel.Aledo@uclm.es}


\begin{abstract}
Bayesian Network (BN) fusion combines multiple input networks into a single structure, balancing dependency preservation with computational tractability. While unrestricted fusion retains all dependencies, it often results in overly complex networks with high treewidth, which affects inference scalability. Limited fusion mitigates this by pruning edges to control treewidth but risks overfitting to input-specific noise and omitting dependencies from the original BNs. This paper introduces a consensus framework that prioritizes shared structures among input networks while enforcing treewidth constraints, ensuring a good consensus. We propose genetic algorithms with advanced initialization, specialized operators, and a tailored fitness function. Additionally, we adapt existing methods to this problem and implement greedy baselines for benchmarking and further optimization. Experiments on synthetic and real-world BNs show the superiority of the proposed genetic algorithms over the adapted methods and greedy baselines.
\end{abstract}

\ccsdesc[500]{Computing methodologies~Genetic algorithms}
\ccsdesc[500]{Mathematics of computing~Bayesian networks}
\ccsdesc[300]{Theory of computation~Evolutionary algorithms}

\keywords{Bayesian Networks, Fusion, Consensus, Genetic Algorithm, Combinatorial Optimization, Federated Learning}

\received{29 January 2025}
\received[revised]{XX XXXX 2025}
\received[accepted]{XX XXXX 2025}

\maketitle

\sloppy

%
%
\section{Introduction} \label{sec:introduction}

Bayesian Networks (BNs) \cite{Jensen_Nielsen,Kjaerulff_Madsen,Koller_Friedman} are probabilistic graphical models that represent complex dependencies between variables through Directed Acyclic Graphs (DAGs). Each BN combines a graph structure encoding conditional (in)dependencies with a set of conditional probability distributions quantifying these dependencies. BNs are highly interpretable due to their graphical structure \cite{relevance_BNs_1997,MeekesRG15} and possess symbolic inference mechanisms \cite{Jensen_Nielsen}, making them effective in fields such as bioinformatics \cite{Angelopoulos_2022,Bernaola2023}, healthcare \cite{McLachlan2020}, environmental modeling \cite{Dai2024}, and industrial diagnostic systems \cite{Adedipe2020,Li2020,Zhao2023}, where they enable inference and decision-making under uncertainty. Moreover, the increasing focus on explainable AI models \cite{xAI:InfFus:2020,Bacardit_GA_XAI_2022} reinforces the relevance of BNs in applications that require transparency due to their white-box nature.

Although domain experts can manually specify the graph structure and parameters of BNs, this approach becomes infeasible for large-scale problems. Learning BNs from data, although NP-hard \cite{learning-BN-NP_2004}, has been extensively studied \cite{chickering_optimal_2002,gamez_learning_2011,Kitson2023,Scanagatta_review_2019,tsamardinos_max-min_2006}. In this context, structural fusion of BNs \cite{pena_finding_2011,Puerta2021Fusion} is crucial when multiple networks, potentially from different experts or data sources, must be merged into a single, coherent model. Given a set of BNs on the same variables, structural fusion aims to create a consensus BN that best represents the input BNs. This is increasingly relevant due to the rise of distributed \cite{Verbraeken2020} and federated learning \cite{Li_review_FL_applications_2020,Zhang_survey_fl_KBS_2021} where BN fusion is essential for aggregating local models, as shown in recent works \cite{Laborda2024_KBS,Laborda2024_IJAR,Torrijos2024_DS}.


Achieving optimal fusion entails constructing a consensus model that captures the conditional dependencies of the input networks. While this problem is NP-hard \cite{pena_finding_2011}, \cite{Puerta2021Fusion} proposed a heuristic approach to approximate the optimal fused BN, significantly reducing computational requirements. However, the standard fusion of BNs often generates dense and complex graphs, limiting its usability in scenarios requiring efficient inference and interpretability.

Inference complexity in a BN depends on its treewidth $t$, defined as the size of the largest subset of nodes fully connected (\textit{clique}) in a triangulated version of the network’s moral graph\footnote{The moral graph of a BN is the undirected graph obtained by linking all pairs of non-adjacent nodes with a common child and then removing edge directions.}. As $t$ increases, inference complexity scales exponentially, $O(n \cdot k^{t+1})$ \cite{Chandrasekaran2008}, where $n$ is the number of variables and $k$ the states per variable, making high treewidth BNs infeasible for practical use. To mitigate this, \cite{Torrijos2024_CEC} proposed a genetic algorithm that prunes edges in the fused network, reducing its treewidth while retaining the essential structure of the unrestricted fusion to enhance scalability and usability.

Given the NP-hard nature of many problems related to BNs, evolutionary computation has been widely employed as a heuristic solution approach. This technique has been utilized not only in structural learning of BNs \cite{Larranaga_review_2013}, but also in triangulating the moral graph needed for BN inference \cite{GamezPuerta_triangulation_02}, and finding the most probable explanation given observations \cite{deCampos1999}.

In this paper, we propose a new fusion methodology designed to achieve a desired treewidth in the final consensus BN by selectively removing edges in the original BNs before fusion using a genetic algorithm \cite{Katoch2020} with operators precisely adjusted to the nature of the problem. We design two distinct approaches: one where each occurrence of an edge in the input BNs is treated as a unique element in the chromosome and can be removed independently, and another where identical edges across the input networks are treated as a single element, with removal decisions applying to all networks simultaneously. For each strategy, a greedy algorithm is developed to provide a baseline for performance comparison.

Moreover, our approach differs from \cite{Torrijos2024_CEC} by redefining the solution quality metric. While the fusion in \cite{Puerta2021Fusion} is theoretically optimal and captures all conditional independencies, such fidelity may not always be desirable. For instance, in a federated learning setting, if one client introduces erroneous or malicious conditional independencies, it could corrupt the fused BN. Thus, our approach minimizes the divergence between the fused BN and all the input networks while achieving target treewidth, improving robustness against inaccurate or adversarial (in)dependencies.

Thus, the main contributions of this work are:
\begin{itemize}
  \item Establishing a new consensus definition for Bayesian Network fusion that minimizes the average divergence between the fused BN and the input networks while complying to a target treewidth constraint.
  \item Proposing two genetic-based algorithms, complemented by corresponding greedy baselines, for achieving BN consensus by selectively pruning edges in the input BNs.
  \item Introducing new metrics to evaluate the quality of fused BNs, focusing on divergence from the input networks.
  \item Providing all the code to allow for reproducibility and boosting research on the topic.
\end{itemize}

The paper is organized as follows: Section \ref{sec:preliminaries} covers the basics of Bayesian Networks. Section \ref{sec:problem_definition} defines the BN consensus problem and its associated constraints and objectives. Section \ref{sec:algorithm} presents the proposed genetic and greedy algorithms for BN consensus under a treewidth constraint. Section \ref{sec:experimental_evaluation} shows experimental results on synthetic and real-world datasets. Section \ref{sec:conclusion} concludes with key findings, implications, and future research directions.

%
%
\section{Preliminaries} \label{sec:preliminaries}

\subsection{Bayesian Networks} \label{subsec:preliminaries_bns}
A Bayesian Network (BN) \cite{Jensen_Nielsen,Kjaerulff_Madsen,Koller_Friedman} is a probabilistic graphical model representing conditional (in)dependencies among a set of variables ${\mathcal{X}} = \{X_1, \dots, X_n\}$. Formally, a BN is defined as a pair $\mathcal{B} = (\mathcal{G}, \mathcal{P})$, where $\mathcal{G} = (\mathcal{X},\mathcal{E})$ is a Directed Acyclic Graph (DAG) with nodes ${\mathcal{X}}$ representing variables and directed edges $\mathcal{E}$ encoding dependencies among these variables, and $\mathcal{P}$ is a set of conditional probability distributions, $\{\mathbb{P}(X_i \mid pa(X_i))\}_{i=1}^n$, where $pa(X_i)$ denotes the parent set of $X_i$ in $\mathcal{G}$.

The structure $\mathcal{G}$ encodes conditional independencies using the criterion of \textit{d-separation} \cite{Koller_Friedman}, which allows factorising the joint probability distribution as $\mathbb{P}(X_1, \dots, X_n) = \prod_{i=1}^n \mathbb{P}(X_i \mid pa(X_i))$. Thus, $\mathcal{G}$ imposes that each variable $X_i$ is conditionally independent of its non-descendants given its parents $pa(X_i)$ in the DAG.

The independencies encoded by $\mathcal{G}$ can be represented as $I(\mathcal{G})$, where each element is a conditional independence relationship of the form $(X_i \perp X_j \mid Z)$, with $X_i$ and $X_j$ conditionally independent given a subset of nodes $Z$. A DAG $\mathcal{G}_1$ is an \textit{I-map} (Independence map) of another DAG $\mathcal{G}_2$ if $I(\mathcal{G}_1) \subseteq I(\mathcal{G}_2)$, meaning that $\mathcal{G}_1$ encodes at least the same conditional independencies as $\mathcal{G}_2$.

A DAG $\mathcal{G}_1$ is a \textit{minimal I-map} of $\mathcal{G}_2$ if it is an $I$-map of $\mathcal{G}_2$ and no arc can be removed from $\mathcal{G}_1$ without violating at least one conditional independence in $\mathcal{G}_2$. Thus, a minimal I-map $\mathcal{G}_1$ is the sparsest graph that retains all conditional independencies of $\mathcal{G}_2$.

\subsection{Structural Fusion of BNs} \label{subsec:fusion_bns}
Structural fusion in Bayesian Networks aims to merge multiple BNs, potentially derived from different data sources or expert knowledge, into a single BN that captures the essential dependencies of the input networks. Let $\mathbb{B} = \{\mathcal{B}_1, \dots, \mathcal{B}_r\}$ be a set of BNs with corresponding DAGs $\mathbb{G} = \{\mathcal{G}_1, \dots, \mathcal{G}_r\}$. The goal is to construct a DAG $\mathcal{G}^+$ that serves as a minimal $I$-map of the intersection of the independence sets in $\mathbb{G}$. Formally, this requires that $I(\mathcal{G}^+) = \bigcap_{i=1}^{r} I(\mathcal{G}_i)$, where ${\mathcal{G}}^+$ minimizes the number of arcs while capturing the shared conditional independencies among all input DAGs. Thus, any conditional independence in $\mathcal{G}^+$ is also present in each $\mathcal{G}_i$.

For example, consider the three input BNs represented by the DAGs in Fig. \ref{fig:ex1-original-bns}, each modeling dependencies among variables $\mathcal{X} = \{X_1, X_2, X_3, X_4\}$. The goal is to produce a fused DAG $\mathcal{G}^+$ that maintains only the conditional independencies common to all input DAGs while minimizing the number of arcs. Fig. \ref{fig:ex2-fusion-bns} illustrates two possible fusion outcomes for the DAGs in Fig. \ref{fig:ex1-original-bns}, where the structure on the right achieves a more optimal fusion by reducing the number of arcs while preserving shared dependencies.

\begin{figure}[htb]
  \centering
          \begin{tikzpicture}[->,>=stealth',shorten >=1pt,auto,node distance=1cm,  
                      semithick]
                \tikzstyle{every state}=[fill=none,draw=black,text=black]
              
                \node[state,inner sep=1.5pt,minimum size=1.5pt]         (X1)                      {$X_1$};
                \node[state,inner sep=1.5pt,minimum size=1.5pt]         (X2) [below of=X1]          {$X_2$};
                \node[state,inner sep=1.5pt,minimum size=1.5pt]         (X3) [right of=X2]         {$X_3$};
                \node[state,inner sep=1.5pt,minimum size=1.5pt]         (X4) [above of=X3]          {$X_4$};
                
                \path (X1) edge [graph2]             node {} (X2)
                      (X2) edge [graph2]             node {} (X3)
                      (X4) edge [graph2]             node {} (X3);
              \end{tikzpicture}
      \hspace{0.5cm}
          \begin{tikzpicture}[->,>=stealth',shorten >=1pt,auto,node distance=1cm,  
                      semithick]
                \tikzstyle{every state}=[fill=none,draw=black,text=black]
              
                \node[state,inner sep=1.5pt,minimum size=1.5pt]         (X1)                      {$X_1$};
                \node[state,inner sep=1.5pt,minimum size=1.5pt]         (X2) [below of=X1]          {$X_2$};
                \node[state,inner sep=1.5pt,minimum size=1.5pt]         (X3) [right of=X2]         {$X_3$};
                \node[state,inner sep=1.5pt,minimum size=1.5pt]         (X4) [above of=X3]          {$X_4$};
                
                \path (X1) edge [graph2]             node {} (X4)
                      (X2) edge [graph2]             node {} (X3)
                      (X4) edge [graph2]             node {} (X3);
              \end{tikzpicture}
      \hspace{0.5cm}
          \begin{tikzpicture}[->,>=stealth',shorten >=1pt,auto,node distance=1cm,  
                      semithick]
                \tikzstyle{every state}=[fill=none,draw=black,text=black]
              
                \node[state,inner sep=1.5pt,minimum size=1.5pt]         (X1)                      {$X_1$};
                \node[state,inner sep=1.5pt,minimum size=1.5pt]         (X2) [below of=X1]          {$X_2$};
                \node[state,inner sep=1.5pt,minimum size=1.5pt]         (X3) [right of=X2]         {$X_3$};
                \node[state,inner sep=1.5pt,minimum size=1.5pt]         (X4) [above of=X3]          {$X_4$};
                
                \path (X1) edge [graph2]             node {} (X4)
                      (X1) edge [graph2]             node {} (X3)
                      (X3) edge [graph2]             node {} (X2);
              \end{tikzpicture}
          \vspace{-0.2cm}  
  \caption{Three input BNs for the fusion process.}
  \Description{Three input BNs with four variables for the fusion process.}
  \label{fig:ex1-original-bns}
\end{figure}
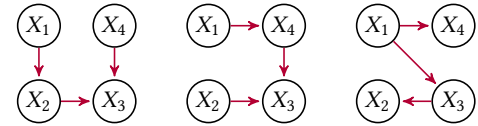

\begin{figure}[htb]
  \centering
          \begin{tikzpicture}[->,>=stealth',shorten >=1pt,auto,node distance=1cm,  
                      semithick]
                \tikzstyle{every state}=[fill=none,draw=black,text=black]
              
                \node[state,inner sep=1.5pt,minimum size=1.5pt]         (X1)                      {$X_1$};
                \node[state,inner sep=1.5pt,minimum size=1.5pt]         (X2) [below of=X1]          {$X_2$};
                \node[state,inner sep=1.5pt,minimum size=1.5pt]         (X3) [right of=X2]         {$X_3$};
                \node[state,inner sep=1.5pt,minimum size=1.5pt]         (X4) [above of=X3]          {$X_4$};
                
                \path (X2) edge [graph2]             node {} (X1)
                      (X3) edge [graph2]             node {} (X1)
                      (X4) edge [graph2]             node {} (X1)
                      (X2) edge [graph2]             node {} (X3)
                      (X2) edge [graph2]             node {} (X4)
                      (X4) edge [graph2]             node {} (X3);
              \end{tikzpicture}
      \hspace{1cm}
          \begin{tikzpicture}[->,>=stealth',shorten >=1pt,auto,node distance=1cm,  
                      semithick]
                \tikzstyle{every state}=[fill=none,draw=black,text=black]
              
                \node[state,inner sep=1.5pt,minimum size=1.5pt]         (X1)                      {$X_1$};
                \node[state,inner sep=1.5pt,minimum size=1.5pt]         (X2) [below of=X1]          {$X_2$};
                \node[state,inner sep=1.5pt,minimum size=1.5pt]         (X3) [right of=X2]         {$X_3$};
                \node[state,inner sep=1.5pt,minimum size=1.5pt]         (X4) [above of=X3]          {$X_4$};
                
                \path (X1) edge [graph2]             node {} (X2)
                      (X1) edge [graph2]             node {} (X3)
                      (X1) edge [graph2]             node {} (X4)
                      (X2) edge [graph2]             node {} (X3)
                      (X4) edge [graph2]             node {} (X3);
              \end{tikzpicture}
           \vspace{-0.2cm}   
  \caption{Two possible BNs resulting from the fusion of the three input networks in Fig. \ref{fig:ex1-original-bns}.}
  \Description{Two possible Bayesian Networks resulting from the fusion of the three input networks in Fig. \ref{fig:ex1-original-bns}.}
  \label{fig:ex2-fusion-bns}
\end{figure}
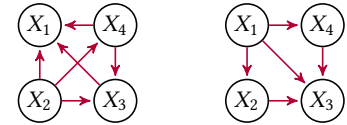

The process for constructing ${\mathcal{G}}^+$ follows these steps \cite{pena_finding_2011,sagrado_qualitative_2003}: (1) Define an ordering $\sigma$ over the variables ${\mathcal{X}}$. (2) Modify each DAG ${\mathcal{G}}_i$ to obtain an minimal $I$-map ${\mathcal{G}}_i^{\sigma}$ of ${\mathcal{G}}_i$ consistent\footnote{A DAG $\mathcal{G}$ over $\mathcal{X}$ is said to be consistent with an ordering $\sigma$ of $\mathcal{X}$ if, for every node $X \in \mathcal{X}$, all parents of $X$ in $\mathcal{G}$ precede $X$ in $\sigma$.} with $\sigma$. (3) Construct ${\mathcal{G}}^+$ as the union of arcs across all ${\mathcal{G}}_i^{\sigma}$, i.e.${\mathcal{G}}^+ = \bigcup_{i=1}^r arcs({\mathcal{G}}^{\sigma}_i)$, where $arcs({\mathcal{G}})$ returns the directed edges in ${\mathcal{G}}$.

\textit{Algorithm A} \cite{pena_finding_2011} is applied to perform step (2), maintaining the conditional independencies of each ${\mathcal{G}}_i$ encoded by $d$-separation. This guarantees that ${\mathcal{G}}^+$ is a minimal $I$-map of the intersection of the independence sets across all input graphs, namely, $I({\mathcal{G}}^+) = \bigcap_{i=1}^r I({\mathcal{G}}_i)$. For the theoretical foundations of algorithm $A$, we refer the reader to \cite{pena_finding_2011,sagrado_qualitative_2003}.

\begin{figure}[htb]
  \centering
          \begin{tikzpicture}[->,>=stealth',shorten >=1pt,auto,node distance=1cm,  
                      semithick]
                \tikzstyle{every state}=[fill=none,draw=black,text=black]
              
                \node[state,inner sep=1.5pt,minimum size=1.5pt]         (X1)                      {$X_1$};
                \node[state,inner sep=1.5pt,minimum size=1.5pt]         (X2) [below of=X1]          {$X_2$};
                \node[state,inner sep=1.5pt,minimum size=1.5pt]         (X3) [right of=X2]         {$X_3$};
                \node[state,inner sep=1.5pt,minimum size=1.5pt]         (X4) [above of=X3]          {$X_4$};
                
                \path (X1) edge [graph2]             node {} (X2)
                      (X2) edge [graph2]             node {} (X3)
                      (X4) edge [graph2]             node {} (X3);
              \end{tikzpicture}
      \hspace{0.5cm}
          \begin{tikzpicture}[->,>=stealth',shorten >=1pt,auto,node distance=1cm,  
                      semithick]
                \tikzstyle{every state}=[fill=none,draw=black,text=black]
              
                \node[state,inner sep=1.5pt,minimum size=1.5pt]         (X1)                      {$X_1$};
                \node[state,inner sep=1.5pt,minimum size=1.5pt]         (X2) [below of=X1]          {$X_2$};
                \node[state,inner sep=1.5pt,minimum size=1.5pt]         (X3) [right of=X2]         {$X_3$};
                \node[state,inner sep=1.5pt,minimum size=1.5pt]         (X4) [above of=X3]          {$X_4$};
                
                \path (X1) edge [graph2]             node {} (X4)
                      (X2) edge [graph2]             node {} (X3)
                      (X4) edge [graph2]             node {} (X3);
              \end{tikzpicture}
      \hspace{0.5cm}
          \begin{tikzpicture}[->,>=stealth',shorten >=1pt,auto,node distance=1cm,  
                      semithick]
                \tikzstyle{every state}=[fill=none,draw=black,text=black]
              
                \node[state,inner sep=1.5pt,minimum size=1.5pt]         (X1)                      {$X_1$};
                \node[state,inner sep=1.5pt,minimum size=1.5pt]         (X2) [below of=X1]          {$X_2$};
                \node[state,inner sep=1.5pt,minimum size=1.5pt]         (X3) [right of=X2]         {$X_3$};
                \node[state,inner sep=1.5pt,minimum size=1.5pt]         (X4) [above of=X3]          {$X_4$};
                
                \path (X1) edge [graph2]             node {} (X2)
                      (X1) edge [graph2]             node {} (X3)
                      (X1) edge [graph2]             node {} (X4)
                      (X2) edge [graph2]             node {} (X3);
              \end{tikzpicture}
          \vspace{-0.2cm}  
  \caption{BNs obtained from those in Fig. \ref{fig:ex1-original-bns} after applying algorithm A with $\sigma = (X_1,X_2,X_4,X_3)$.}
  \Description{BNs obtained from those in Fig. \ref{fig:ex1-original-bns} after applying algorithm A with $\sigma = (X_1,X_2,X_4,X_3)$.}
  \label{fig:ex3-transformed-bns}
\end{figure}
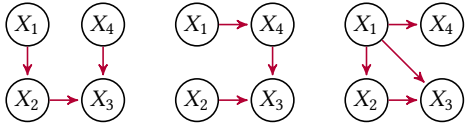

The choice of the variable ordering $\sigma$ is critical in determining the structure of ${\mathcal{G}}^+$. Algorithm $A$ applies this ordering to each input DAG ${\mathcal{G}}_i$, yielding the minimal $I$-map ${\mathcal{G}}_i^{\sigma}$ of ${\mathcal{G}}_i$ consistent with $\sigma$. For example, Fig. \ref{fig:ex3-transformed-bns} illustrates the transformation of DAGs from Fig. \ref{fig:ex1-original-bns} using the ordering $\sigma = (X_1, X_2, X_4, X_3)$.
This ordering $\sigma$ determines the fused BN shown in Fig. \ref{fig:ex2-fusion-bns} (right).

Although finding the optimal ordering $\sigma$ is an NP-hard problem, practical heuristics \cite{Puerta2021Fusion} offer efficient approximate solutions that yield near-optimal fused networks. This is crucial, as a poorly chosen $\sigma$ can result in unnecessarily complex fusions.

The fusion method proposed in \cite{Puerta2021Fusion} captures all conditional dependencies across the input DAGs $\mathbb{G} = \{\mathcal{G}_1, \dots, \mathcal{G}_r\}$, but it imposes an ``all-or-nothing'' requirement: if a dependency exists in any single network ${\mathcal{G}}_i$, it must also appear in the fused structure $\mathcal{G}^+$, regardless of how prevalent it is among the input networks. This strict criterion often results in dense networks with very high treewidth, which increases the computational burden of parameter estimation and inference. High treewidth can render the fused network impractical for applications. To mitigate this issue, we propose exploring a more flexible fusion approach to produce a consensus network with lower treewidth, focusing on capturing a representative subset of dependencies from the input graphs rather than enforcing all conditional dependencies.

\subsection{Metrics for Measuring Structural Similarity of BNs} \label{subsec:preliminaries_metrics}
Several metrics can be used to evaluate the structural similarity of BNs. The Structural Hamming Distance (SHD) \cite{tsamardinos_max-min_2006} quantifies structural differences by the minimum number of arc modifications (additions, deletions, or reversals) needed to align two BN structures. However, it neglects conditional independencies encoded by the networks, potentially misrepresenting similarity \cite{Puerta2021Fusion}.

To address this, the Structural Moral Hamming Distance (SMHD) \cite{Kim2019SHDMoral,Laborda2024_IJAR,Laborda2024_KBS,Torrijos2024_CEC} compares the moral graphs of two networks 
\begin{equation*}
    \text{SMHD}(\mathcal{G}_1, \mathcal{G}_2) = \left| \mathcal{E}_{\text{moral}}(\mathcal{G}_1) \triangle \mathcal{E}_{\text{moral}}(\mathcal{G}_2) \right|,
\end{equation*}
where $\mathcal{E}_{\text{moral}}(\mathcal{G})$ is the set of undirected edges in the moralized graph of $\mathcal{G}$, and $\triangle$ represents the symmetric difference.

Complementing this, the Fusion Similarity (\textsc{FSim}, introduced in \cite{Puerta2021Fusion}, Section 6.2) evaluates the structural similarity based on the minimal $I$-maps of the networks under a shared topological order $\sigma$ of the variables. Unlike SMHD, which focuses on the structural differences of the graphs, \textsc{FSim} emphasizes the differences in the conditional independencies encoded by the BNs.




%
%
\section{Problem Definition} \label{sec:problem_definition}

The fusion method proposed in \cite{Puerta2021Fusion} generates a fused network $\mathcal{G}^+$ that retains all dependencies present in any of the input networks $\mathbb{G} = \{\mathcal{G}_1, \dots, \mathcal{G}_r\}$. While theoretically optimal in preserving all conditional dependencies of the input networks, this approach often yields a network with high treewidth. To address this, \cite{Torrijos2024_CEC} introduced the concept of \textit{Restricted Structural Fusion of BNs}, where edges in the fused network $\mathcal{G}^+$ are selectively removed to reduce the treewidth under a limit while minimizing the Structural Moral Hamming Distance (SMHD) to $\mathcal{G}^+$. This approach aims to approximate $\mathcal{G}^+$ closely while constraining the treewidth, resulting in a more tractable network denoted as $\mathcal{G}^+_{tw \leq t}$.

However, using $\mathcal{G}^+$ as the reference structure has limitations. Since $\mathcal{G}^+$ aggregates all dependences present in the input BNs, it may introduce unnecessary complexity. This issue is particularly evident when one or more input networks contain noisy or irrelevant dependencies, which can disproportionately distort the fusion when $\mathcal{G}^+$ serves as the reference.

\begin{figure}[htb]
  \centering
      \subfloat[$\mathcal{G}_1$. \label{subfig:ex1-BN1}]{%
          \begin{tikzpicture}[->,>=stealth',shorten >=1pt,auto,node distance=1cm,  
                      semithick]
                \tikzstyle{every state}=[fill=none,draw=black,text=black]

                \node[state,inner sep=1.5pt,minimum size=1.5pt]         (X1)                      {$X_2$};
                \node[state,inner sep=1.5pt,minimum size=1.5pt]         (X4) [below left of=X1]          {$X_1$};
                \node[state,inner sep=1.5pt,minimum size=1.5pt]         (X3) [below right of=X1]         {$X_3$};
                \node[state,inner sep=1.5pt,minimum size=1.5pt]         (X2) [below of=X4]          {$X_5$};
                \node[state,inner sep=1.5pt,minimum size=1.5pt]         (X5) [below of=X3]          {$X_4$};
                
                \path (X1) edge [graph2]             node {} (X4)
                      (X1) edge [graph2]             node {} (X3)
                      (X4) edge [graph2]             node {} (X2)
                      (X3) edge [graph2]             node {} (X5);

              \end{tikzpicture}}
      \hspace{0.8cm}
      \subfloat[$\mathcal{G}_2$. \label{subfig:ex1-BN2}]{%
          \begin{tikzpicture}[->,>=stealth',shorten >=1pt,auto,node distance=1cm,
                      semithick]
                \tikzstyle{every state}=[fill=none,draw=black,text=black]
              
                \node[state,inner sep=1.5pt,minimum size=1.5pt]         (X1)                      {$X_1$};
                \node[state,inner sep=1.5pt,minimum size=1.5pt]         (X4) [below left of=X1]          {$X_3$};
                \node[state,inner sep=1.5pt,minimum size=1.5pt]         (X3) [below right of=X1]         {$X_4$};
                \node[state,inner sep=1.5pt,minimum size=1.5pt]         (X2) [below of=X4]          {$X_5$};
                \node[state,inner sep=1.5pt,minimum size=1.5pt]         (X5) [below of=X3]          {$X_2$};
                
                \path (X1) edge [graph2]             node {} (X4)
                      (X1) edge [graph2]             node {} (X3)
                      (X4) edge [graph2]             node {} (X2)
                      (X3) edge [graph2]             node {} (X5);
              \end{tikzpicture}}
      \hspace{0.8cm}
      \subfloat[$\mathcal{G}_3$. \label{subfig:ex1-BN3}]{%
          \begin{tikzpicture}[->,>=stealth',shorten >=1pt,auto,node distance=1cm,
                      semithick]
                \tikzstyle{every state}=[fill=none,draw=black,text=black]
              
                \node[state,inner sep=1.5pt,minimum size=1.5pt]         (X1)                      {$X_3$};
                \node[state,inner sep=1.5pt,minimum size=1.5pt]         (X4) [below left of=X1]          {$X_1$};
                \node[state,inner sep=1.5pt,minimum size=1.5pt]         (X3) [below right of=X1]         {$X_5$};
                \node[state,inner sep=1.5pt,minimum size=1.5pt]         (X2) [below of=X4]          {$X_4$};
                \node[state,inner sep=1.5pt,minimum size=1.5pt]         (X5) [below of=X3]          {$X_2$};
                
                \path (X1) edge [graph2]             node {} (X4)
                      (X1) edge [graph2]             node {} (X3)
                      (X4) edge [graph2]             node {} (X2)
                      (X3) edge [graph2]             node {} (X5);
              \end{tikzpicture}}
      \\
      \vspace{-0.3cm}
      \subfloat[$\mathcal{G}^+$. \label{subfig:ex1-BNfusion}]{%
          \begin{tikzpicture}[->,>=stealth',shorten >=1pt,auto,node distance=0.9cm,
                      semithick]
                \tikzstyle{every state}=[fill=none,draw=black,text=black]
              
                \node[state,inner sep=1.5pt,minimum size=1.5pt]         (X4)           {$X_4$};
                \node[state,inner sep=1.5pt,minimum size=1.5pt]         (X2) [below left of=X4]         {$X_2$};
                \node[state,inner sep=1.5pt,minimum size=1.5pt]         (X3) [below right of=X4]          {$X_3$};
                \node[state,inner sep=1.5pt,minimum size=1.5pt]         (X5) [below right of=X2]          {$X_5$};
                \node[state,inner sep=1.5pt,minimum size=1.5pt]         (X1) [above right of=X3]                {$X_1$};
                
                \path (X1) edge [graph2, bend right=75]             node {} (X2)
                      (X1) edge [graph2]             node {} (X4)
                      (X1) edge [graph2]             node {} (X3)
                      (X1) edge [graph2, bend left=50]             node {} (X5)

                      (X3) edge [graph2]             node {} (X4)
                      (X3) edge [graph2]             node {} (X5)
                      (X3) edge [graph2]             node {} (X2)

                      (X4) edge [graph2]             node {} (X2)
                      (X5) edge [graph2]             node {} (X2);
              \end{tikzpicture}}
      \hspace{0.3cm}
      \subfloat[$\mathcal{G}^+_{tw\leq3}$. \label{subfig:ex1-BNfusionLimitedG+}]{%
          \begin{tikzpicture}[->,>=stealth',shorten >=1pt,auto,node distance=0.9cm,
                      semithick]
                \tikzstyle{every state}=[fill=none,draw=black,text=black]
              
                \node[state,inner sep=1.5pt,minimum size=1.5pt]         (X4)           {$X_4$};
                \node[state,inner sep=1.5pt,minimum size=1.5pt]         (X2) [below left of=X4]         {$X_2$};
                \node[state,inner sep=1.5pt,minimum size=1.5pt]         (X3) [below right of=X4]          {$X_3$};
                \node[state,inner sep=1.5pt,minimum size=1.5pt]         (X5) [below right of=X2]          {$X_5$};
                \node[state,inner sep=1.5pt,minimum size=1.5pt]         (X1) [above right of=X3]                {$X_1$};
                
                \path (X1) edge [graph2, bend right=75]             node {} (X2)
                      (X1) edge [graph2]             node {} (X4)
                      (X1) edge [graph2]             node {} (X3)
                      (X1) edge [graph2, bend left=50]             node {} (X5)

                      (X3) edge [graph2]             node {} (X4)
                      (X3) edge [graph2]             node {} (X5)
                      
                      (X4) edge [graph2]             node {} (X2);
              \end{tikzpicture}}
      \hspace{0.3cm}
      \subfloat[$\mathcal{G}^*_{tw\leq3}$. \label{subfig:ex1-BNfusionLimitedG*}]{%
          \begin{tikzpicture}[->,>=stealth',shorten >=1pt,auto,node distance=0.9cm,
                      semithick]
                \tikzstyle{every state}=[fill=none,draw=black,text=black]
              
                \node[state,inner sep=1.5pt,minimum size=1.5pt]         (X4)           {$X_4$};
                \node[state,inner sep=1.5pt,minimum size=1.5pt]         (X2) [below left of=X4]         {$X_2$};
                \node[state,inner sep=1.5pt,minimum size=1.5pt]         (X3) [below right of=X4]          {$X_3$};
                \node[state,inner sep=1.5pt,minimum size=1.5pt]         (X5) [below right of=X2]          {$X_5$};
                \node[state,inner sep=1.5pt,minimum size=1.5pt]         (X1) [above right of=X3]                {$X_1$};
                
                \path (X1) edge [graph2]             node {} (X4)
                      (X1) edge [graph2]             node {} (X3)

                      (X3) edge [graph2]             node {} (X5);
              \end{tikzpicture}}
  \caption{Fusion of $\{\mathcal{G}_1,\mathcal{G}_2,\mathcal{G}_3\}$: unrestricted ($\mathcal{G}^+$) and restricted to $tw=3$ ($\mathcal{G}^+_{tw\leq3}$ and $\mathcal{G}^*_{tw\leq3}$).}
  \Description{Fusion of $\{\mathcal{G}_1,\mathcal{G}_2,\mathcal{G}_3\}$: unrestricted ($\mathcal{G}^+$) and restricted to $tw=3$ ($\mathcal{G}^+_{tw\leq3}$ and $\mathcal{G}^*_{tw\leq3}$).}
  \label{fig:ex1-treewidth}
\end{figure}
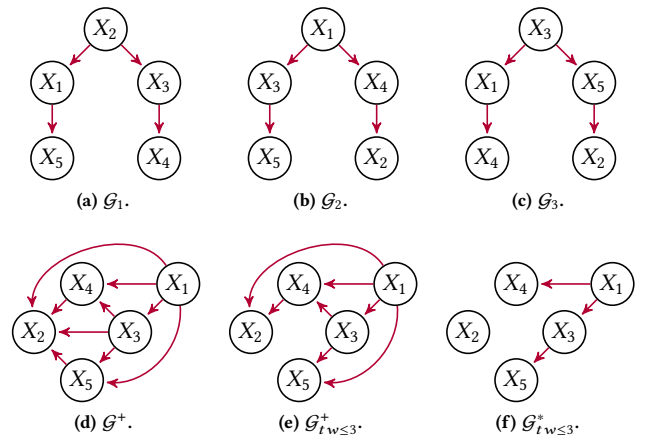

Consider the example in Fig. \ref{fig:ex1-treewidth} (with metric scores in Table \ref{tab:ex1-ex2-metrics}). Here, three input networks (each with treewidth 2) are fused (Figs. \ref{subfig:ex1-BN1}, \ref{subfig:ex1-BN2}, and \ref{subfig:ex1-BN3}). The unrestricted fusion $\mathcal{G}^+$ (Fig. \ref{subfig:ex1-BNfusion}) results in a network with treewidth 5. In contrast, the constrained treewidth approach in \cite{Torrijos2024_CEC} yields $\mathcal{G}^+_{tw \leq 3}$ (Fig. \ref{subfig:ex1-BNfusionLimitedG+}), which approximates $\mathcal{G}^+$ by removing only two edges, resulting in a complex network but achieving treewidth 3. Note that Fig. \ref{subfig:ex1-BNfusionLimitedG*} is described later in the document for a different example after Definition \ref{def:problem_new}.

\begin{table}[htb]
  \caption{Metrics for BNs in Figure \ref{fig:ex1-treewidth} and Figure \ref{fig:ex2-treewidth}. 
           \textsc{tw} denotes the treewidth of the BN; 
           $\text{SMHD}_{\mathcal{G}^+}$ is the SMHD with respect to the fusion $\mathcal{G}^+$; 
           \SMHD~ is the average SMHD over the initial BNs $\mathbb{G}$; 
           and \FSim~ is the average \textsc{FSim} over $\mathbb{G}$.  Dashes indicate cases where the metric is not meaningful, such as when averages would involve self-comparisons.}
  \label{tab:ex1-ex2-metrics}
  \resizebox{\columnwidth}{!} {%
    \begin{tabular*}{1.5\columnwidth}{@{\extracolsep{\fill}}@{\hspace{0.1em}}l@{\hspace{0.5em}}S[table-format=1.0]@{\hspace{0.5em}}S[table-format=1.0]@{\hspace{0.5em}}S[table-format=1.3]@{\hspace{0.5em}}S[table-format=1.3]@{\hspace{1em}}l@{\hspace{0.5em}}S[table-format=1.0]@{\hspace{0.5em}}S[table-format=1.0]@{\hspace{0.5em}}S[table-format=1.3]@{\hspace{0.5em}}S[table-format=1.3]@{\hspace{0.1em}}}
    \toprule
    \multicolumn{1}{c}{\multirow{2}{*}{\textsc{\bfseries BN}}}
    & \multicolumn{4}{c}{\textsc{\bfseries Figure \ref{fig:ex1-treewidth}}} 
    & \multicolumn{1}{c}{\multirow{2}{*}{\textsc{\bfseries BN}}}
    & \multicolumn{4}{c}{\textsc{\bfseries Figure \ref{fig:ex2-treewidth}}} \\
    \cmidrule(r){2-5} \cmidrule(){7-10}
    & \textsc{tw} & \textsc{$\text{SMHD}_{\mathcal{G}^+}$} & \textsc{SMHD$_{\mathbb{G}}$} & \textsc{FSim$_{\mathbb{G}}$} 
    & & \textsc{tw} & \textsc{$\text{SMHD}_{\mathcal{G}^+}$} & \textsc{SMHD$_{\mathbb{G}}$} & \textsc{FSim$_{\mathbb{G}}$} \\
    \cmidrule[0.8pt](r){1-5}\cmidrule[0.9pt](){6-10}
    $\{\mathcal{G}_i\}_{i=1}^{3}$   & 2 & 6 & {---} & {---} &      $\{\mathcal{G}_i\}_{i=1}^{99}$  & 2 & 6   & {---} & {---} \\
                                     & & & & &                      $\mathcal{G}_{100}$             & 5 & 0 & {---} & {---} \\
    $\mathcal{G}^+$                  & 5 & {---} & 6 & 7.33 &       $\mathcal{G}^+$                 & 5 & {---} & 5.94 & 4.95 \\
    $\mathcal{G}^+_{tw\leq3}$        & 3 & \B 3 & 4.33 & 5.33 &     $\mathcal{G}^+_{tw\leq3}$       & 3 & \B 3 & 3 & 2.99 \\
    $\mathcal{G}^*_{tw\leq3}$        & 2 & 7 & \B 3 & \B 3.33 &     $\mathcal{G}^*_{tw\leq3}$       & 2 & 6 & \B 0.06 & \B 0.05 \\
    \bottomrule
    \end{tabular*}
  }
\end{table}

\begin{figure}[htb]
  \centering
      \subfloat[$\{\mathcal{G}_i\}_{i=1}^{99},\mathcal{G}^*_{tw\leq 3}$. \label{subfig:ex2-BNi}]{%
          \begin{tikzpicture}[->,>=stealth',shorten >=1pt,auto,node distance=1cm,
                      semithick]
                \tikzstyle{every state}=[fill=none,draw=black,text=black]
              
                \node[state,inner sep=1.5pt,minimum size=1.5pt]         (X1)                      {$X_1$};
                \node[state,inner sep=1.5pt,minimum size=1.5pt]         (X4) [below left of=X1]          {$X_3$};
                \node[state,inner sep=1.5pt,minimum size=1.5pt]         (X3) [below right of=X1]         {$X_4$};
                \node[state,inner sep=1.5pt,minimum size=1.5pt]         (X2) [below of=X4]          {$X_5$};
                \node[state,inner sep=1.5pt,minimum size=1.5pt]         (X5) [below of=X3]          {$X_2$};
                
                \path (X1) edge [graph2]             node {} (X4)
                      (X1) edge [graph2]             node {} (X3)
                      (X4) edge [graph2]             node {} (X2)
                      (X3) edge [graph2]             node {} (X5);
              \end{tikzpicture}}
      \hspace{0.2cm}
      \subfloat[$\mathcal{G}_{100},\mathcal{G}^+$. \label{subfig:ex2-BN100}]{%
          \begin{tikzpicture}[->,>=stealth',shorten >=1pt,auto,node distance=1cm,
                      semithick]
                \tikzstyle{every state}=[fill=none,draw=black,text=black]
              
                \node[state,inner sep=1.5pt,minimum size=1.5pt]         (X4)           {$X_4$};
                \node[state,inner sep=1.5pt,minimum size=1.5pt]         (X2) [below left of=X4]         {$X_2$};
                \node[state,inner sep=1.5pt,minimum size=1.5pt]         (X3) [below right of=X4]          {$X_3$};
                \node[state,inner sep=1.5pt,minimum size=1.5pt]         (X5) [below right of=X2]          {$X_5$};
                \node[state,inner sep=1.5pt,minimum size=1.5pt]         (X1) [above right of=X3]                {$X_1$};
                
                \path (X1) edge [graph2, bend right=75]             node {} (X2)
                      (X1) edge [graph2]             node {} (X4)
                      (X1) edge [graph2]             node {} (X3)
                      (X1) edge [graph2, bend left=50]             node {} (X5)

                      (X3) edge [graph2]             node {} (X4)
                      (X3) edge [graph2]             node {} (X5)
                      (X3) edge [graph2]             node {} (X2)

                      (X4) edge [graph2]             node {} (X2)
                      (X5) edge [graph2]             node {} (X2);
              \end{tikzpicture}}
      \hspace{0.2cm}
      \subfloat[$\mathcal{G}^+_{tw\leq3}$. \label{subfig:ex2-BNfusionLimitedG+}]{%
          \begin{tikzpicture}[->,>=stealth',shorten >=1pt,auto,node distance=0.9cm,
                      semithick]
                \tikzstyle{every state}=[fill=none,draw=black,text=black]
              
                \node[state,inner sep=1.5pt,minimum size=1.5pt]         (X4)           {$X_4$};
                \node[state,inner sep=1.5pt,minimum size=1.5pt]         (X2) [below left of=X4]         {$X_2$};
                \node[state,inner sep=1.5pt,minimum size=1.5pt]         (X3) [below right of=X4]          {$X_3$};
                \node[state,inner sep=1.5pt,minimum size=1.5pt]         (X5) [below right of=X2]          {$X_5$};
                \node[state,inner sep=1.5pt,minimum size=1.5pt]         (X1) [above right of=X3]                {$X_1$};
                
                \path (X1) edge [graph2, bend right=75]             node {} (X2)
                      (X1) edge [graph2]             node {} (X4)
                      (X1) edge [graph2]             node {} (X3)
                      (X1) edge [graph2, bend left=50]             node {} (X5)

                      (X3) edge [graph2]             node {} (X4)
                      (X3) edge [graph2]             node {} (X5)
                      
                      (X4) edge [graph2]             node {} (X2);
              \end{tikzpicture}}
  \caption{Fusion of $\{\mathcal{G}_i\}_{i=1}^{100}$: unrestricted ($\mathcal{G}^+$) and restricted to $tw=3$ ($\mathcal{G}^+_{tw\leq3}$ and $\mathcal{G}^*_{tw\leq3}$).}
  \Description{Fusion of $\{\mathcal{G}_i\}_{i=1}^{100}$: unrestricted ($\mathcal{G}^+$) and restricted to $tw=3$ ($\mathcal{G}^+_{tw\leq3}$ and $\mathcal{G}^*_{tw\leq3}$).}
  \label{fig:ex2-treewidth}
\end{figure}
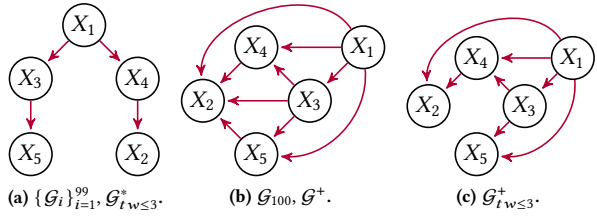

In a second example (Fig. \ref{fig:ex2-treewidth}, with metric scores in Table \ref{tab:ex1-ex2-metrics}), the effect of noise on the fusion process is more pronounced. Here, 99 input networks with treewidth 2 share dependencies (Fig. \ref{subfig:ex2-BNi}), but a single input network $\mathcal{G}_{100}$ (Fig. \ref{subfig:ex2-BN100}) introduces spurious dependencies. This could, for example, be a case of federated learning in which a client is malicious. As a result, the unrestricted fusion $\mathcal{G}^+$ incorporates these noisy dependencies and becomes overly dense, essentially replicating $\mathcal{G}_{100}$. Even with treewidth constrained, $\mathcal{G}^+_{tw \leq 3}$ (Fig. \ref{subfig:ex2-BNfusionLimitedG+}) fails to represent a consensus of the majority due to the excessive influence of $\mathcal{G}_{100}$.

These examples illustrate that defining restricted fusion based on $\mathcal{G}^+$ can lead to a complex fused network that fails to capture the (in)dependencies of the input BNs. In response, we propose a new problem definition that avoids using $\mathcal{G}^+$ as a reference. Instead, we directly measure similarity to each input network $\mathcal{G}_i$, defining the fused network as one that minimizes the average structural distance to all input networks under a treewidth constraint, thus better capturing shared dependencies and mitigating noise or errors.

\begin{definition}[Restricted Structural Consensus of Bayesian Networks Relative to Original Graphs]\label{def:problem_new}
    Let $\mathbb{G} = \{{\mathcal{G}}_1, \dots, {\mathcal{G}}_r\}$ be the DAGs representing the graphical structure of $r$ BNs defined over the same domain. The goal of the restricted structural consensus relative to the input graphs is to obtain a fused graph $\mathcal{G}^*_{tw \leq t}$, which maximizes the average similarity between the fused graph ${\mathcal{G}}$ and the input graphs ${\mathcal{G}}_i$, subject to a treewidth constraint, by minimizing a defined structural distance metric. Specifically, for a given maximum treewidth $t \in \mathds{N}$, with $t \geq 2$, the restricted structural fusion is defined as
    \begin{equation}
        \mathcal{G}^*_{tw \leq t} = \arg \min_{\mathcal{G}; \,\textrm{tw}(\mathcal{G}) \leq t} \frac{1}{r} \sum_{i=1}^{r} Metric\, (\mathcal{G}, \mathcal{G}_i)
    \label{eq:problem_new}
    \end{equation}
    where $\textrm{Metric}\ ({\mathcal{G}}, {\mathcal{G}}_i)$ represents a structural distance metric, such as SMHD or \textsc{FSim}, used to evaluate the similarity between the fused graph ${\mathcal{G}}$ and each of the original networks ${\mathcal{G}}_i$.
\end{definition}

This definition relaxes the notion of \textit{fusion} to a \textit{consensus} in which the resulting network $\mathcal{G}^*_{tw\leq t}$ captures essential dependencies shared across input networks, subject to a treewidth constraint.

Revisiting the examples, in Figure \ref{fig:ex1-treewidth}, while the input networks have simple structures, their conditional independencies vary, making the unrestricted fusion $\mathcal{G}^+$ complex. Our consensus approach (Fig. \ref{subfig:ex1-BNfusionLimitedG*}) generates a $\mathcal{G}^*_{tw\leq 3}$ network with fewer edges than even the input networks, reflecting a more balanced consensus. Metrics in Table \ref{tab:ex1-ex2-metrics} show that $\mathcal{G}^+_{tw\leq 3}$ achieves better SMHD$_\mathcal{G^+}$, while $\mathcal{G}^*_{tw\leq 3}$ achieves better \SMHD~ and \FSim~.

This effect is even clearer in the second example (Fig. \ref{fig:ex2-treewidth}). Since $\mathcal{G}^+$ and $\mathcal{G}^+_{tw \leq 3}$ (Figs. \ref{subfig:ex2-BN100} and \ref{subfig:ex2-BNfusionLimitedG+}) both over-represent the noisy dependencies from $\mathcal{G}_{100}$, our consensus definition $\mathcal{G}^*_{tw\leq 3}$ (Fig. \ref{subfig:ex2-BN100}) reflects the shared structure of $\{\mathcal{G}_i\}_{i=1}^{99}$ while omitting the noise introducing by $\mathcal{G}_{100}$. As shown in Table \ref{tab:ex1-ex2-metrics}, the new consensus approach yields significantly better \SMHD~ and \FSim~, nearly matching the ideal structure shared by most input networks.

An additional advantage of this consensus approach is that unlike the restricted fusion method from \cite{Torrijos2024_CEC}, which tends to exploit the maximum allowable treewidth $t$ to minimize SMHD$_\mathcal{G^+}$, our new approach produces $\mathcal{G}^*_{tw\leq t}$ consensus BNs that maintain manageable treewidth values, often similar to or even lower than those of the input BNs. Consequently, our method ensures tractable networks with lower complexity, facilitating efficient parametric learning and inference operations.

%
%
\section{Genetic Approach for BN Consensus Under Treewidth Constraints} \label{sec:algorithm}
This work presents a genetic algorithm for the consensus of Bayesian networks introduced in Definition \ref{def:problem_new} under a treewidth constraint ($t$). The approach prunes arcs from the input networks to construct a consensus structure that balances structural similarity and computational feasibility. By integrating a problem-specific greedy initialization with tailored genetic operators, the algorithm ensures efficient exploration and exploitation of the search space while adhering to the treewidth limitation.

\subsection{Chromosome Representation} \label{subsec:algorithm_chromosome}

The representation of a chromosome $C$ depends on the choice of the edge set $\mathcal{E}$, with two possible configurations: 

\paragraph{Without Repetition: \Ec.} $C$ represents all unique edges across the input graphs: $\mathcal{E}_{\mathbb{G}} = \bigcup_{i=1}^r \text{edges}(\mathcal{G}_i)$. Let $s = |\mathcal{E}_{\mathbb{G}}|$, with edges arranged in lexicographical order by their subscripts (e.g., $a_{1,3} \prec a_{1,6} \prec a_{3,4}$). A chromosome $C \in \{0,1\}^s$ encodes whether each edge is included (1) or excluded (0) across all input networks.

\paragraph{With Repetition: \Eb.} $C$ accounts for all edges from each input graph separately: $\mathcal{E}_{\mathbb{G}}^* = \bigsqcup_{i=1}^r \text{edges}(\mathcal{G}_i)$. Let $s^* = |\mathcal{E}_{\mathbb{G}}^*|$, with edges arranged in lexicographical order by their subscripts and the graph to which they belong (e.g., $a_{2,3}^1 \prec a_{3,4}^1 \prec a_{1,2}^3$). A chromosome $C \in \{0,1\}^{s^*}$ specifies the inclusion (1) or exclusion (0) of each edge for each graph independently.

\paragraph{Trade-off.} The choice between \Eb~ and \Ec~ balances simplicity and flexibility. \Ec~ defines a smaller search space, facilitating faster convergence and consistent pruning across networks but limiting specificity. In contrast, \Eb~ allows network-specific pruning, potentially achieving better solutions but requiring exploration of a larger and more computationally intensive search space.

\subsection{Fitness Function} \label{subsec:algorithm_fitness}

The fitness function evaluates a chromosome $C$ based on the structural similarity of the fused graph $\mathcal{G}^*_C$ to the original input networks $\{\mathcal{G}_1, \dots, \mathcal{G}_r\}$, while penalizing violations of the treewidth constraint.

Given $C$, the modified input graphs $\{\mathcal{G}_C^1, \dots, \mathcal{G}_C^r\}$ are defined as:
\begin{equation*}
    \mathcal{G}_C^i = (\mathcal{X}, \mathcal{E}_C^i); \; \mathcal{E}_C^i = 
    \begin{cases} 
        \{a_{jk}^i \in \text{edges}(\mathcal{G}_i) \mid C(a_{jk}^i) = 1\}, & \text{if } \mathcal{E} = \mathcal{E}_{\mathbb{G}} \\
        \{a_{jk} \in \text{edges}(\mathcal{G}_i) \mid C(a_{jk}) = 1\}, & \text{if } \mathcal{E} = \mathcal{E}_{\mathbb{G}}^*
    \end{cases}
\end{equation*}

The fused graph $\mathcal{G}^*_C$ is obtained by applying the method in Section \ref{subsec:fusion_bns} \cite{Puerta2021Fusion} to $\{\mathcal{G}_C^1, \dots, \mathcal{G}_C^r\}$. The fitness is then computed as:
\begin{equation*}
    f(C) = \frac{1}{r} \sum_{i=1}^r \text{Metric}(\mathcal{G}^*_C, \mathcal{G}_i) \cdot 
    \begin{cases} 
        1, & \text{if } \text{tw}(\mathcal{G}^*_C) \leq t, \\
        \frac{\text{tw}(\mathcal{G}^*_C)}{t} & \text{otherwise,}
    \end{cases}
\end{equation*}
where $\text{tw}(\mathcal{G})$ denotes the treewidth of $\mathcal{G}$. Minimizing $f(C)$ ensures high structural similarity to the input graphs while preserving the treewidth constraint. The penalty term guides the search away from infeasible solutions.

\subsection{Algorithm Structure} \label{subsec:algorithm_structure}

The proposed genetic algorithm, detailed in Alg. \ref{alg:genetic}, tackles the challenge of BN consensus under a treewidth constraint (Definition \ref{def:problem_new}). 
The decoding of chromosomes into graphs and the computation of the fitness function are computationally demanding, requiring a careful trade-off between solution quality and efficiency.

\begin{algorithm}[htb]
\caption{Genetic Pre-Fusion Edge Pruning for BN Consensus with Limited Treewidth}
\label{alg:genetic}
\begin{algorithmic}[1] 
    \Require $\{{\mathcal{G}}_1, \dots,{\mathcal{G}}_r\}$ defined over ${\mathcal{X}}$; $t \geq 2$; $nIterations$; $popSize$; $rep$ (boolean)
    \Ensure $\{{\mathcal{G}}_1^*, \dots,{\mathcal{G}}_r^*\}, {\mathcal{G}}^*_{tw \leq t}$

        \State $\mathcal{E} \leftarrow \left\{
            \begin{array}{ll}
            \bigsqcup_{i=1}^r \textrm{edges}(\mathcal{G}_i) & \text{if } rep \\ 
            \bigcup_{i=1}^r \textrm{edges}(\mathcal{G}_i) & \text{o.w.}
            \end{array}
            \right.$

        \State $\left\{
            \begin{array}{ll}
                \#(a_{jk}^i) \leftarrow \sum_{i=1}^r {\mathds{1}}(a_{jk}^i \in \mathcal{G}_i), \; \forall a_{jk}^i \in \mathcal{E} & \text{if } rep \\
                \#(a_{jk}) \leftarrow \sum_{i=1}^r {\mathds{1}}(a_{jk} \in \mathcal{G}_i), \; \forall a_{jk} \in \mathcal{E} & \text{o.w.}
            \end{array}
            \right.$
            
        \State $\{{\mathcal{G}}_1^*, \dots,{\mathcal{G}}_r^*\} \leftarrow \textrm{empty graphs over} \; {\mathcal{X}}$
        \State $C^*, C^l \leftarrow \emptyset$  \Comment{Best chromosome seen so far/in the last iteration}

        \State $\mathcal{C}_{gr} \leftarrow \{\textrm{Alg. \ref{alg:greedy}}(\{{\mathcal{G}}_1, \dots,{\mathcal{G}}_r\}, t, rep)\}$
        \State $\mathcal{C}_{gr} \leftarrow \mathcal{C}_{gr} \cup \{\textrm{Alg. \ref{alg:greedy}}(\{{\mathcal{G}}_1, \dots,{\mathcal{G}}_r\}, t - 1, rep)\} \; \text{if } t > 2$
        \State $\mathcal{C} \leftarrow \textrm{initialization}(popSize - |\mathcal{C}_{gr}|, \#(a_{jk}), \mathcal{E}) \cup \mathcal{C}_{gr}$

        \For{$i \leftarrow 1 \textrm{ to } nIterations$} 
            \State $C^*, C^l, \{{\mathcal{G}}_1^*, \dots,{\mathcal{G}}_r^*\}, {\mathcal{G}}^*_{tw \leq t} \leftarrow$ \textrm{evaluate}($\mathcal{C},\{{\mathcal{G}}_1, \dots,{\mathcal{G}}_r\})$
            \State $\mathcal{C} \leftarrow \textrm{selection}(\mathcal{C})$
            \State $\mathcal{C} \leftarrow \textrm{crossover}(\mathcal{C})$ 
            \State $\mathcal{C} \leftarrow \textrm{mutation}(\mathcal{C})$
            \State $\mathcal{C} \leftarrow (\mathcal{C} \setminus \{C_{worst_1},C_{worst_2}\})\cup \{C^*,C^l\}$ \Comment{Elitism}
        \EndFor
        \State Return $\{{\mathcal{G}}_1^*, \dots,{\mathcal{G}}_r^*\}, {\mathcal{G}}^*_{tw \leq t}$
    \end{algorithmic}
\end{algorithm}

Due to the exponential growth of the search space with the size of the candidate edge set $|\mathcal{E}|$, the algorithm employs a fixed-iteration, population-based approach. Excessive population sizes or iterations are avoided to manage the complexity of evaluating individuals, instead focusing on efficient initialization and tailored operators to maximize iteration effectiveness. Key features of the algorithm include:
\begin{itemize}
    \item \textbf{Hybrid initialization}: Combines heuristic solutions from the greedy algorithm (Alg. \ref{alg:greedy}) with random chromosomes, mixing quality starting points and population diversity.
    \item \textbf{Problem-specific operators}: Mutation is tailored to the search space's structured nature, focusing on efficiently producing high-quality offspring.
    \item \textbf{Constraint-guided fitness}: Solutions exceeding the treewidth constraint are penalized, steering the population toward feasible configurations.
\end{itemize}

The following subsections provide details on initialization, crossover, mutation, and population update strategies.

\subsubsection{Initialization} \label{subsubsec:algorithm_initialization}
The population initialization (size $popSize$) combines heuristic solutions from the problem-specific \textit{greedy pre-fusion edge pruning algorithm} (Alg. \ref{alg:greedy}) with probabilistic sampling to ensure both quality and diversity. The greedy algorithm, tailored for this task, iteratively adds edges to maximize structural similarity while respecting treewidth constraints, with variants for handling edge repetitions. Two chromosomes are generated using this method: one for treewidth $t$ and another for $t - 1$ (if $t > 2$).

\begin{algorithm}[htb]
\caption{Greedy Pre-Fusion Edge Pruning for BN Consensus with Limited Treewidth}
\label{alg:greedy}
\begin{algorithmic}[1] 
    \Require $\{{\mathcal{G}}_1, \dots,{\mathcal{G}}_r\}$ defined over ${\mathcal{X}}$; $t \geq 2$; $rep$ (boolean)
    \Ensure $\{{\mathcal{G}}_1^*, \dots,{\mathcal{G}}_r^*\}, {\mathcal{G}}^*_{tw \leq t}$

        \State $\mathcal{E} \leftarrow \left\{
            \begin{array}{ll}
                \bigsqcup_{i=1}^r \textrm{edges}(\mathcal{G}_i) & \text{if } rep \\ 
                \bigcup_{i=1}^r \textrm{edges}(\mathcal{G}_i) & \text{o.w.}
            \end{array}
            \right.$

        \State $\left\{
            \begin{array}{ll}
                \#(a_{jk}^i) \leftarrow \sum_{i=1}^r {\mathds{1}}(a_{jk}^i \in \mathcal{G}_i), \; \forall a_{jk}^i \in \mathcal{E} & \text{if } rep \\
                \#(a_{jk}) \leftarrow \sum_{i=1}^r {\mathds{1}}(a_{jk} \in \mathcal{G}_i), \; \forall a_{jk} \in \mathcal{E} & \text{o.w.}
            \end{array}
            \right.$

        \State $\{{\mathcal{G}}_1^*, \dots,{\mathcal{G}}_r^*\} \leftarrow \textrm{empty graphs over} \; {\mathcal{X}}$
        \State ${\mathcal{G}}^*_{tw \leq t} \leftarrow \emptyset$

        \While {$\mathcal{E} \neq \emptyset$}
            \State $a_{jk} \gets \arg\max_{a_{jk} \in \mathcal{E}} \#(a_{jk})$
            \State $\mathcal{E} \gets \mathcal{E} \setminus \{a_{jk}\}$
            \State $\mathcal{G}_i^* \leftarrow \left\{
                \begin{array}{ll}
                \mathcal{G}_i^* \cup \{a_{jk}^i\}, \forall i :a_{jk}^i \in \text{edges}(\mathcal{G}_i) & \text{if } rep \\ 
                \mathcal{G}_i^* \cup \{a_{jk}\}, \text{for } \exists! i : a_{jk} \in \text{edges}(\mathcal{G}_i) & \text{o.w.}
                \end{array}
                \right.$

            \State $\sigma \leftarrow$ an ordering for ${\mathcal{X}}$ \Comment Use \cite{Puerta2021Fusion}
            
            \State ${\mathcal{G}}^{\sigma}_i \leftarrow A({\mathcal{G}}_i^*, \sigma) \; \forall i \in \{1, \dots, r\}$

            \If{$\textrm{tw}(\bigcup_{i=1}^r {\mathcal{G}}^{\sigma}_i) \leq t$}
                \State ${\mathcal{G}}^*_{tw \leq t} \leftarrow \bigcup_{i=1}^r {\mathcal{G}}^{\sigma}_i$
            \Else
                \State $\mathcal{G}_i^* \leftarrow \left\{
                    \begin{array}{ll}
                    \mathcal{G}_i^* \setminus \{a_{jk}^i\}, \forall i : a_{jk}^i \in \text{edges}(\mathcal{G}_i) & \text{if } rep \\ 
                    \mathcal{G}_i^* \setminus \{a_{jk}\}, \text{for } \exists! i : a_{jk} \in \text{edges}(\mathcal{G}_i) & \text{o.w.}
                    \end{array}
                    \right.$
            \EndIf
        \EndWhile
        
        \State Return $\{{\mathcal{G}}_1^*, \dots,{\mathcal{G}}_r^*\}, {\mathcal{G}}^*_{tw \leq t}$

    \end{algorithmic}
\end{algorithm}

The remaining chromosomes are initialized probabilistically, guided by the edge frequencies $\#(a_{jk})$ in the input graphs\footnote{By abuse of notation, $a_{jk}$ refers to both $a_{jk}$ and $a_{jk}^i$ in this paragraph.}. For each edge $a_{jk} \in \mathcal{E}$, its inclusion probability is $\mathbb{P}(C(a_{jk}) = 1) = \frac{1}{1 - \log(x)}$, where $x$ is the min-max normalized value\footnote{Uniform random initialization ($\mathbb{P} = 0.5$) occurs when all edges have equal frequency.} of $\#(a_{jk})$.

\subsubsection{Crossover} \label{subsec:algorithm_crossover}
Crossover operates by pairing the selected individuals (with a tournament selection of size 2) and applying a common single-point crossover. A random crossover point is chosen for each pair, splitting their chromosomes into two parts. The offspring are generated by combining the first segment of one parent with the second segment of the other, resulting in $popSize$ new individuals in the population.

\subsubsection{Mutation} \label{subsec:algorithm_mutation}
In the mutation stage, all chromosomes are candidates for potential modifications. The probabilities for adding or removing edges are dynamically adjusted based on the ratio $x = \frac{\text{tw}(\mathcal{G}^*_C)}{t}$, balancing exploration and exploitation. The mutation probabilities are defined as
\begin{equation*}
    \mathbb{P}(\text{add}), \mathbb{P}(\text{remove}) =
    \begin{cases}
        \left(\frac{x-1}{2}\right)^2 + 0.01, \quad \frac{x}{10}, & x \leq 1 \\
        \frac{0.01}{x}, \quad \frac{\log(x)}{3} + 0.01, & x > 1
    \end{cases}.
\end{equation*}

Each edge in the chromosome is mutated independently. For edge addition, a candidate edge $a_{jk} \notin \mathcal{E}_C$ is added with probability $\mathbb{P}(\text{add})$. For edge removal, an existing edge $a_{jk} \in \mathcal{E}_C$ is removed with probability $\mathbb{P}(\text{remove})$. This adaptive approach ensures a balanced search space exploration while steering solutions toward the feasible region with treewidth $\leq t$.

\subsubsection{Population Update} \label{subsec:algorithm_population_update}
After the mutation step, the population is updated using an elitist strategy to balance genetic diversity and solution quality. Specifically, the two least fit individuals in the population are replaced by (1) $C^*$, the best chromosome found during the evolutionary process, and (2) $C^l$, the best chromosome from the previous generation. This approach ensures that high-quality genetic material persists across generations while maintaining diversity to prevent premature convergence.

%
%
\section{Experimental Evaluation} \label{sec:experimental_evaluation}
To ensure comparability, our experimental setup follows the approach in \cite{Puerta2021Fusion,Torrijos2024_CEC}, extended to use larger Bayesian Networks (BNs) and new metrics relevant to this problem.

\subsection{Networks / datasets} \label{subsec:exp_networks}
As in \cite{Torrijos2024_CEC}, we evaluate using synthetic and real-world BNs. Input DAGs $\mathbb{G} = \{\mathcal{G}_1,\dots,\mathcal{G}_r\}$ are generated from a base DAG $\mathcal{G}_0$ by applying the method in \cite{Puerta2021Fusion,Torrijos2024_CEC}. Each network of $\mathbb{G}$ undergoes $p = n \cdot 0.75$ perturbations, where $n$ is the number of nodes. A randomly selected edge $\{X \rightarrow Y\}$ is either added or removed in each perturbation, ensuring acyclicity. BN complexity is constrained by limiting each node to three parents, four children, and $e = n \cdot 2.5$ edges in the network, preserving structural similarity to $\mathcal{G}_0$. 

For the synthetic networks, $\mathcal{G}_0$ is generated using the method proposed in \cite{Melanon2004}, as in \cite{Puerta2021Fusion,Torrijos2024_CEC}. For real-world networks, we employ seven BNs from the \texttt{bnlearn} Bayesian Network Repository\footnote{\url{https://www.bnlearn.com/bnrepository/}}. These networks, selected from various domains, represent a range of complexities regarding node count, edge density, and inference parameters, as seen in Table \ref{tab:used_BNs}.

To facilitate parametric learning on the generated BNs in assessing their potential use for inference across different treewidth limits, we also generate samples of 5000 instances for each real-world BN using the corresponding DAG and parameters.

\begin{table}[htb]
  \caption{Real-world BNs used in the experiments.}
  \label{tab:used_BNs}
  \resizebox{\columnwidth}{!} {%
    \begin{tabular*}{1.35\columnwidth}{@{\extracolsep{\fill}}lS[table-format=3.0]S[table-format=4.0]S[table-format=5.0]S[table-format=6.0]S[table-format=3.0]S[table-format=6.0]S[table-format=4.0]}
    \toprule
    & \textsc{Child} & \textsc{Insurance} & \textsc{Water} & \textsc{Mildew} & \textsc{Alarm} & \textsc{Barley} & \textsc{Hailfinder} \\
    \midrule
    \textsc{\#Nodes}       & 20 & 27 & 32 & 35 & 37 & 48 & 56  \\
    \textsc{\#Edges}       & 25 & 52 & 66 & 46 & 46 & 84 & 66  \\
    \textsc{\#Params}       & 230 & 1008 & 10083 & 540150 & 509 & 114005 & 2656  \\
    \bottomrule
    \end{tabular*}
  }
\end{table}

\subsection{Algorithms} \label{subsec:algorithms}
This study evaluates the performance of genetic and greedy algorithms for BN consensus under a treewidth constraint, considering various candidate edge set configurations:

\begin{itemize} 
    \item \textbf{\Ea:} Edges are derived from the fused network $\mathcal{G}^+$, as defined in \cite{Torrijos2024_CEC}. The greedy and genetic algorithms from \cite{Torrijos2024_CEC} are adapted to optimize metrics relative to the original input graphs rather than $\mathcal{G}^+$.

    \item \textbf{\Eb:} Edges include repetitions from each input graph, \Eb$ = \bigsqcup_{i=1}^r \textrm{edges}(\mathcal{G}_i)$. The proposed genetic algorithm (Alg. \ref{alg:genetic}) and its greedy baseline (Alg. \ref{alg:greedy}) are applied to allow independent inclusion decisions (\texttt{rep = True}). 
    
    \item \textbf{\Ec:} Edges are unique across the input graphs, \Ec$ = \bigcup_{i=1}^r \textrm{edges}(\mathcal{G}_i)$. Alg. \ref{alg:genetic} and Alg. \ref{alg:greedy} are applied with uniform inclusion decisions (\texttt{rep = False}).
\end{itemize}

This setup allows us to investigate how the adapted algorithm from \cite{Torrijos2024_CEC} performs in the new consensus framework and whether changing edges in the original graphs before fusion improves the quality of the constrained consensus BN.

\subsection{Reproducibility} \label{subsec:reproducibility}
All algorithms were implemented in Java (OpenJDK 17) using the Tetrad 7.6.5 causal reasoning library\footnote{\url{https://github.com/cmu-phil/tetrad/releases/tag/v7.6.5}}. For reproducibility, the complete code, BNs, and datasets are available on GitHub\footnote{\url{https://github.com/ptorrijos99/BNFusion}}, with datasets also hosted in OpenML\footnote{\url{https://www.openml.org/search?type=data&tags.tag=bnlearn}}. All experiments were conducted in a controlled environment on Rocky Linux 8.9 machines with AMD EPYC 7453 processors, using seven threads and 8 GB RAM per run.

\subsection{Methodology} \label{subsec:exp_methodology}
To scale up from prior work \cite{Torrijos2024_CEC}, we evaluate larger networks with more input DAGs. For both synthetic and real-world BNs, we set $r = \{10, 30, 50\}$ as the number of input DAGs. We vary the number of nodes for synthetic networks with $n = \{10, 30, 50\}$. Real-world networks, in contrast, have a fixed number of nodes ranging from 20 in \texttt{child} to 56 in \texttt{hailfinder}, as detailed in Table \ref{tab:used_BNs}.

To assess the fusion quality, we introduce two metrics (Section \ref{subsec:preliminaries_metrics}): the average SMHD score and Fusion Similarity relative to the initial set of DAGs, $\mathbb{G} = \{\mathcal{G}_1, \dots, \mathcal{G}_r\}$. We denote these metrics as $m = \{\textsc{SMHD}_{\mathbb{G}}, \textsc{FSim}_{\mathbb{G}}\}$.

For each base DAG $\mathcal{G}_0$, ten sets of input DAGs $\mathbb{G}^i = \{\mathcal{G}_1^i,\dots,\mathcal{G}_r^i\}$ were generated using distinct random seeds, along with ten datasets $\mathbb{D}^i$ of 5000 instances for real-world networks.
For each $(i, r, n, m, tw)$ combination, with $tw \in \{t \mid 2 \leq t < \text{tw}(\mathcal{G}^+)\}$, each genetic algorithm and its greedy counterpart from Section \ref{subsec:algorithms} was run. Genetic algorithms used a population size of 100, following the optimal configuration identified in \cite{Torrijos2024_CEC}.

Parametric learning was conducted using the Expectation Maximization (EM) method implemented in the Tetrad library. The gold-standard network was reparametrized with the dataset generated from the corresponding random seed.

Results were analyzed using the methodology in \cite{exreport-fuente1, exreport-fuente2} with the \texttt{exreport} R package \cite{exreport}. A Friedman test \cite{Friedman1940} was applied to test the null hypothesis $H_0$ that all algorithms are equal. If rejected, a post-hoc Holm procedure \cite{Holm1979} compared the algorithms to the top-ranked one, with both tests at a 1\% significance level.

\subsection{Results for Synthetic BNs} \label{subsec:results_synthetic_BNs}
\begin{figure}[htbp]
  \centering
  \includegraphics[width=\columnwidth]{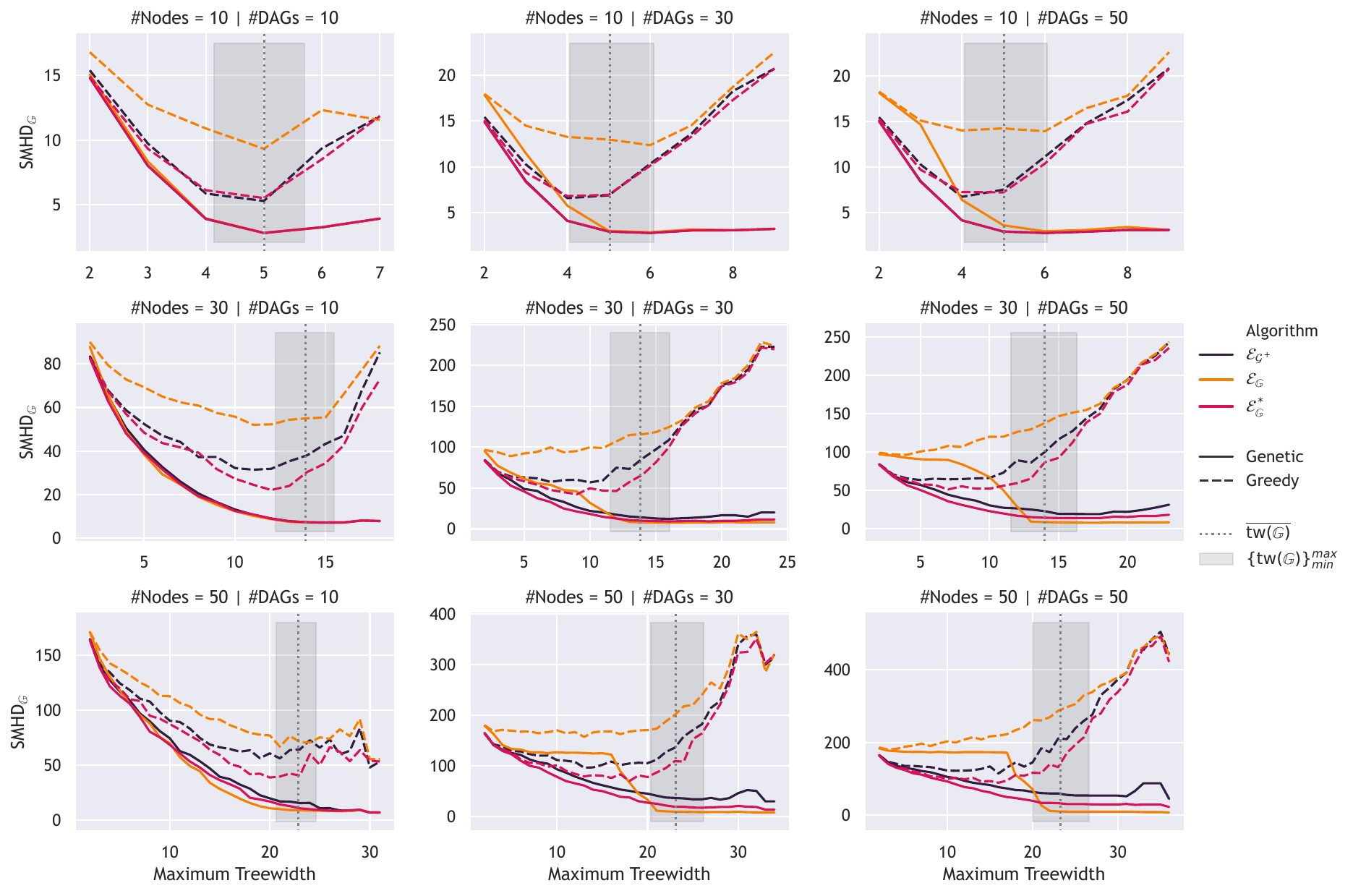}
  \Description{\SMHD~ in synthetic BNs, for each possible $t$.}
  \caption{\SMHD~ in synthetic BNs, for each possible $t$.}
  \label{fig:synthetic_SMHD_original}
\end{figure}

\begin{figure}[htbp]
  \centering
  \includegraphics[width=\columnwidth]{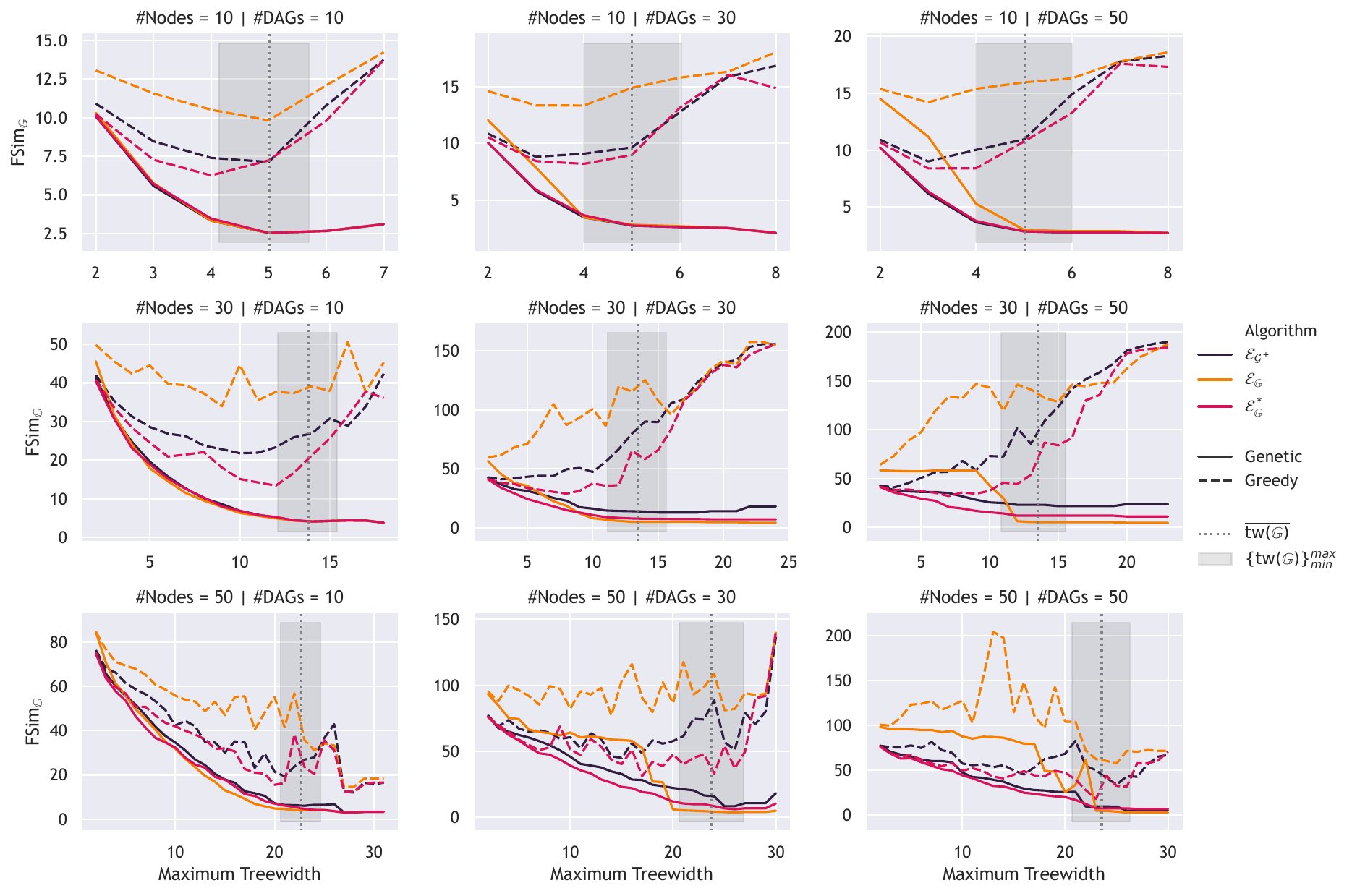}
  \Description{\FSim~ in synthetic BNs, for each possible $t$.}
  \caption{\FSim~ in synthetic BNs, for each possible $t$.}
  \label{fig:synthetic_FSim_original}
\end{figure}

This section evaluates the proposed algorithms on synthetic BNs, aiming to minimize two key metrics: SMHD and \textsc{FSim} (Section \ref{subsec:preliminaries_metrics}) relative to the original input networks, \SMHD~ and \FSim, respectively. These metrics measure the fused network’s structural similarity to the input BNs. Two independent runs were conducted for the genetic algorithms: one optimizing \SMHD~ and the other \FSim. Figures \ref{fig:synthetic_SMHD_original} and \ref{fig:synthetic_FSim_original} summarize the results, showing average performance across various treewidth constraints, along with the range $\{\text{tw}(\mathbb{G})\}_{min}^{max}$ and average $\overline{\text{tw}(\mathbb{G})}$ of the input graphs for context.

Table \ref{tab:synthetic_results} summarizes the results, including the mean difference between the best-performing algorithm and others ($\overline{\textsc{Diff}}$), the number of best performances (\textsc{\#Best}), average execution time ($\overline{\textsc{Time (s)}}$), and the \textsc{$p$-value} from Holm's post-hoc test. The Friedman test yields $1.05 \times 10^{-141}$ for \SMHD~ and $1.72 \times 10^{-128}$ for \FSim, rejecting $H_0$ in both cases. For context, the unrestricted $\mathcal{G}^+$ fusion method shows a mean difference of 211.5 in \SMHD~ and 98.9 in \FSim~ from the best-performing algorithm across all datasets.

\begin{table}[htbp]
  \caption{Results on synthetic BNs.}
  \label{tab:synthetic_results}
  \resizebox{\columnwidth}{!} {%
    \begin{tabular*}{1.35\columnwidth}{@{\extracolsep{\fill}}@{\hspace{0.1em}}l@{\hspace{0.5em}}l@{\hspace{0.5em}}S[table-format=3.1]@{\hspace{0.5em}}S[table-format=1.2e-2]@{\hspace{0.1em}}S[table-format=3.0]@{\hspace{0.5em}}S[table-format=3.1]@{\hspace{0.5em}}S[table-format=2.1]@{\hspace{0.5em}}S[table-format=1.2e-2]@{\hspace{0.1em}}S[table-format=3.0]@{\hspace{0.5em}}S[table-format=4.1]@{\extracolsep{\fill}}@{\hspace{0.1em}}}
      \toprule
      \multicolumn{2}{c}{\multirowcell{2}{\textsc{\bfseries Algorithm}}} & \multicolumn{4}{c}{\textsc{\bfseries SMHD$_\mathbb{G}$}} & \multicolumn{4}{c}{\textsc{\bfseries FSim$_\mathbb{G}$}} \\
      \cmidrule(r){3-6}\cmidrule(){7-10}
      & & $\overline{\textsc{Diff}}$ & \textsc{$p$-value} & \textsc{\#Best} & $\overline{\textsc{Time (s)}}$ & $\overline{\textsc{Diff}}$ & \textsc{$p$-value} & \textsc{\#Best} & $\overline{\textsc{Time (s)}}$ \\
      \midrule
      \multirowcell{3}{\rotatebox[origin=c]{90}{\textsc{Genetic}}} 
        & \Ec~ (Alg. \ref{alg:genetic}) & \B 3.5 & {---} & \B 95 & 50.5 & \B 1.4 & {---} & 90 & 1790.5 \\
        & \Eb~ (Alg. \ref{alg:genetic}) & 15.1 & 9.16e-07 & \B 95 & 250.0 & 9.5 & 1.22e-05 & \B 91 & 2386.2 \\
        & \Ea~ \cite{Torrijos2024_CEC} & 12.6 & 1.34e-06 & 27 & 4.1 & 5.4 & 2.31e-05 & 30 & 1785.9 \\
      \cmidrule(){1-2}\cmidrule(r){3-6}\cmidrule(){7-10}
      \multirowcell{3}{\rotatebox[origin=c]{90}{\textsc{Greedy}}} 
        & \Ec~ (Alg. \ref{alg:greedy}) & 73.0 & 4.38e-25 & 1 & 8.5 & 30.9 & 4.23e-24 & 1 & 14.3 \\
        & \Eb~ (Alg. \ref{alg:greedy}) & 114.5 & 4.07e-108 & 0 & 127.8 & 62.7 & 3.57e-96 & 0 & 274.6 \\
        & \Ea~ \cite{Torrijos2024_CEC} & 85.8 & 6.24e-56 & 0 & \B 0.1 & 39.7 & 2.86e-50 & 0 & \B 0.1 \\
      \midrule
    \end{tabular*}
  }
\end{table}

Based on these results, we can draw the following conclusions:

\paragraph{Alg. \ref{alg:genetic} vs. \cite{Torrijos2024_CEC}} Alg. \ref{alg:genetic} without edge repetition (\Ec) consistently achieves the best results across all configurations, offering stability in both \SMHD~ and \FSim, though with slightly longer runtimes than the genetic algorithm in \cite{Torrijos2024_CEC} using \Ea. With edge repetition (\Eb), Alg. \ref{alg:genetic} excels at higher treewidths but struggles in constrained settings due to an expanded search space. However, such scenarios, where fusion is restricted to treewidths smaller than those of the original graphs, are uncommon and may reduce the practical value of the fusion process. Statistical tests confirm significant differences among all approaches in both \SMHD~ and \FSim, rejecting $H_0$, with \Ec~ emerging as the best-performing method.

\paragraph{Greedy Algorithms} The greedy Algorithm \ref{alg:greedy} with edge repetition (\Eb) performs poorly in all settings, with high runtimes due to a larger search space. This also negatively impacts its corresponding genetic algorithm, leading to suboptimal results and increased runtimes. However, genetic algorithms effectively mitigate these limitations, demonstrating strong adaptability, as the greedy methods serve only as initialization and are not the focus of this work.

\paragraph{\SMHD~ vs. \FSim} Results for \SMHD~ and \FSim~ are highly correlated, with differences mainly in execution time. \FSim~ requires repeated fusions, making it computationally heavier and reducing runtime disparities across configurations.

\subsection{Results on Real-World BNs} \label{subsec:results_real_BNs}
This section evaluates the algorithms on real-world BNs, focusing on structural accuracy (\SMHD, \FSim) and parametric estimation.

\subsubsection{Structural Accuracy of the Generated Model}
Table \ref{tab:real_results} summarizes the mean structural accuracy of the generated models for the real-world BNs described in Table \ref{tab:used_BNs}. The metrics are the same as in Table \ref{tab:synthetic_results}. The Friedman test yields $1.12 \times 10^{-150}$ for \SMHD~ and $2.93 \times 10^{-130}$ for \FSim, rejecting $H_0$ in both cases.

\begin{table}[htbp]
  \caption{Results on real-world BNs.}
  \label{tab:real_results}
  \resizebox{\columnwidth}{!} {%
    \begin{tabular*}{1.35\columnwidth}{@{\extracolsep{\fill}}@{\hspace{0.1em}}l@{\hspace{0.5em}}l@{\hspace{0.5em}}S[table-format=3.1]@{\hspace{0.5em}}S[table-format=1.2e-2]@{\hspace{0.1em}}S[table-format=3.0]@{\hspace{0.5em}}S[table-format=3.1]@{\hspace{0.5em}}S[table-format=2.1]@{\hspace{0.5em}}S[table-format=1.2e-2]@{\hspace{0.1em}}S[table-format=3.0]@{\hspace{0.5em}}S[table-format=4.1]@{\extracolsep{\fill}}@{\hspace{0.1em}}}
      \toprule
      \multicolumn{2}{c}{\multirowcell{2}{\textsc{\bfseries Algorithm}}} & \multicolumn{4}{c}{\textsc{\bfseries SMHD$_\mathbb{G}$}} & \multicolumn{4}{c}{\textsc{\bfseries FSim$_\mathbb{G}$}} \\
      \cmidrule(r){3-6}\cmidrule(){7-10}
      & & $\overline{\textsc{Diff}}$ & \textsc{$p$-value} & \textsc{\#Best} & $\overline{\textsc{Time (s)}}$ & $\overline{\textsc{Diff}}$ & \textsc{$p$-value} & \textsc{\#Best} & $\overline{\textsc{Time (s)}}$ \\
      \midrule
      \multirowcell{3}{\rotatebox[origin=c]{90}{\textsc{Genetic}}} 
      & \Ec~ (Alg. \ref{alg:genetic}) & 12.6 & {---} & 176 & 160.8 & \B 4.0 & {---} & 178 & 1596.1 \\
      & \Eb~ (Alg. \ref{alg:genetic}) & \B 9.3 & \B 1.57e-02 & \B 379 & 397.1 & 7.2 & \B 2.01e-02 & \B 278 & 1970.5 \\
      & \Ea~ \cite{Torrijos2024_CEC} & 22.5 & 1.46e-08 & 27 & 9.2 & 11.2 & 4.46e-07 & 30 & 1703.0 \\
      \cmidrule(){1-2}\cmidrule(r){3-6}\cmidrule(){7-10}
      \multirowcell{3}{\rotatebox[origin=c]{90}{\textsc{Greedy}}} 
      & \Ec~ (Alg. \ref{alg:greedy}) & 160.4 & 1.25e-34 & 2 & 18.4 & 51.5 & 2.14e-24 & 1 & 13.0 \\
      & \Eb~ (Alg. \ref{alg:greedy}) & 192.3 & 1.07e-86 & 0 & 167.5 & 96.5 & 2.56e-82 & 0 & 128.2 \\
      & \Ea~ \cite{Torrijos2024_CEC} & 179.5 & 7.01e-78 & 0 & \B 0.2 & 71.1 & 1.12e-61 & 0 & \B 0.2 \\
      \midrule
    \end{tabular*}
  }
\end{table}

Real-world results align with synthetic ones, with some notable differences. \Eb~ achieves the most wins in both \SMHD~ and \FSim~ and the best average in \SMHD, while \Ec~ leads on average in \FSim. Both genetic methods outperform \cite{Torrijos2024_CEC}, which remains faster for \SMHD. The gap between greedy and genetic algorithms is even more pronounced, reinforcing the robustness of the genetic approach. Statistical tests confirm \Ec~ as the best method, though not significantly different from \Eb, while both significantly outperform all others.

\subsubsection{Parameter Estimation in the Generated Model} 
This section evaluates the parametric learnability of the generated BNs by measuring average learning time ($\overline{\textsc{Time}^*}$), mean absolute error ($\overline{\textsc{Abs Err}^*}$) against gold-standard computed marginal probabilities, and the frequency of learning failures ($\textsc{\#NaN}$), where a failure is counted if any seed results in failure (maximum value for $\textsc{\#NaN}$ is 551). $\overline{\textsc{Time}^*}$ and $\overline{\textsc{Abs Err}^*}$ are only computed when all algorithms finish successfully, i.e., if $\mathcal{G}^+$ completes execution. The results, summarized in Table \ref{tab:real_inference}, compare the parametric estimation performance on $\mathcal{G}^+$ and the average of the input networks $\overline{\mathbb{G}}$. The table also reports Holm post-hoc test p-values for mean absolute errors. The Friedman test yields $2.24 \times 10^{-106}$ for \SMHD~ and $1.43 \times 10^{-100}$ for \FSim, rejecting $H_0$ in both cases.

\begin{table}[htbp]
  \caption{Parametric estimation results on real-world BNs.}
  \label{tab:real_inference}
  \resizebox{\columnwidth}{!} {%
    \begin{tabular*}{1.42\columnwidth}{@{\extracolsep{\fill}}@{\hspace{0.1em}}l@{\hspace{0.5em}}l@{\hspace{0.5em}}S[table-format=3.2]@{\hspace{0.5em}}S[table-format=1.4]@{\hspace{0.5em}}S[table-format=1.2e-2]@{\hspace{0.1em}}S[table-format=3.0]@{\hspace{0.5em}}S[table-format=3.2]@{\hspace{0.5em}}S[table-format=1.4]@{\hspace{0.5em}}S[table-format=1.2e-2]@{\hspace{0.1em}}S[table-format=3.0]@{\extracolsep{\fill}}@{\hspace{0.1em}}}
      \toprule
      \multicolumn{2}{c}{\multirowcell{2}{\textsc{\bfseries Algorithm}}} & \multicolumn{4}{c}{\textsc{\bfseries SMHD$_\mathbb{G}$}} & \multicolumn{4}{c}{\textsc{\bfseries FSim$_\mathbb{G}$}} \\
      \cmidrule(r){3-6}\cmidrule(){7-10}
      & & $\overline{\textsc{Time}^*}$ & $\overline{\textsc{Abs Err}^*}$ & \textsc{$p$-value} & $\textsc{\#NaN}$ & $\overline{\textsc{Time}^*}$ & $\overline{\textsc{Abs Err}^*}$ & \textsc{$p$-value} & $\textsc{\#NaN}$ \\
      \midrule
      & $\overline{\mathbb{G}}$ & 1.32 & \B 0.0381 & {---} & \B 0 & 2.09 & \B 0.0385 & {---} & \B 0 \\
      \cmidrule(){1-2}\cmidrule(r){3-6}\cmidrule(){7-10}
      \multirowcell{3}{\rotatebox[origin=c]{90}{\textsc{Genetic}}} 
      & \Ec~ (Alg. \ref{alg:genetic}) & 1.02 & 0.0412 & 5.76e-07 & 4 & \B 0.95 & 0.0410 & \B 7.21e-02 & 2 \\
      & \Eb~ (Alg. \ref{alg:genetic}) & \B 0.93 & 0.0406 & \B 3.02e-02 & 7 & 0.98 & 0.0414 & 6.44e-04 & 2 \\
      & \Ea~ \cite{Torrijos2024_CEC} & 1.02 & 0.0421 & 1.74e-13 & 7 & 1.05 & 0.0417 & 4.09e-04 & 2 \\
      \cmidrule(){1-2}\cmidrule(r){3-6}\cmidrule(){7-10}
      \multirowcell{3}{\rotatebox[origin=c]{90}{\textsc{Greedy}}} 
      & \Ec~ (Alg. \ref{alg:greedy}) & 18.01 & 0.0466 & 1.90e-14 & 123 & 17.99 & 0.0478 & 2.78e-14 & 43 \\
      & \Eb~ (Alg. \ref{alg:greedy}) & 22.17 & 0.0498 & 4.36e-53 & 116 & 26.05 & 0.0515 & 1.31e-47 & 58 \\
      & \Ea~ \cite{Torrijos2024_CEC} & 19.05 & 0.0485 & 3.35e-25 & 146 & 23.23 & 0.0497 & 3.43e-23 & 56 \\
      \cmidrule(){1-2}\cmidrule(r){3-6}\cmidrule(){7-10}
      & $\mathcal{G}^+$ & 112.84 & 0.0554 & 8.05e-67 & 280 & 145.03 & 0.0568 & 3.87e-55 & 280 \\
      \midrule
    \end{tabular*}
  }
\end{table}

The results align with expectations. The cases where conditional probability distributions cannot be estimated ($\textsc{\#NaN}$) grow exponentially for algorithms producing the densest networks. The original graphs $\overline{\mathbb{G}}$ achieve the lowest error, while the unrestricted fusion $\mathcal{G}^+$ shows the highest. Greedy algorithms closely approximate $\mathcal{G}^+$, while genetic algorithms perform better, generating networks similar or simpler than the originals with reduced execution times. Failures are fewer than with greedy algorithms, due to memory limitations as the treewidth constraint approaches that of $\mathcal{G}^+$. Both genetic algorithms outperform the one from \cite{Torrijos2024_CEC}, with minimal differences between those optimized with \SMHD~ or \FSim. Statistical tests confirm the original graphs $\overline{\mathbb{G}}$ as the best method, with no significant difference from \Eb~ in \SMHD~ and \Ec~ in \FSim.

%
%

%
%
\section{Conclusions} \label{sec:conclusion}
This study proposed a novel framework for Bayesian Network (BN) fusion, redefining the task as achieving structural consensus under a treewidth constraint. Traditional methods \cite{Puerta2021Fusion} often generate overly complex networks, while constrained approaches \cite{Torrijos2024_CEC} risk overfitting to input noise. Our consensus-driven approach addresses these issues by emphasizing shared structures and minimizing divergence between the fused network and the input BNs. To solve this problem, we designed two specialized genetic algorithms featuring advanced initialization strategies, problem-specific genetic operators, and a fitness function that balances structural similarity with treewidth constraints. Furthermore, we introduced metrics to evaluate consensus quality while controlling BN complexity.

Experimental results on synthetic and real-world datasets demonstrated the effectiveness of the proposed methods. The genetic algorithms consistently outperformed an adapted version of prior work \cite{Torrijos2024_CEC}, confirming the benefits of pruning edges directly in the input BNs before fusion. Furthermore, the practical usability of the fused networks was validated by examining the learnability of their conditional probability tables (CPTs). Notably, the algorithm allowing edge repetitions while exploring a larger and more complex search space achieved superior solutions in specific cases, underscoring the trade-off between convergence stability and solution quality.

In conclusion, the proposed genetic algorithms offer a robust and flexible solution for constrained BN consensus. Future research could focus on developing enhanced greedy heuristics for initialization, exploring advanced optimization strategies, and extending the framework to applications such as federated learning and other domains requiring constrained graphical model integration.

\begin{acks}
This work is funded by 
\grantsponsor{MICIU}{MICIU/AEI/10.13039/501100011033}{https://doi.org/10.13039/501100011033} and \grantsponsor{PRTR}{European Union NextGenerationEU/PRTR}{} under project \grantnum{TED2021}{TED2021-131291B-I00};
\grantsponsor{JCCM}{Junta de Comunidades de Castilla-La Mancha}{} and \grantsponsor{ERDF_E}{ERDF, EU}{} under project \grantnum{Junta}{SBPLY/21/180225/000062};
\grantsponsor{MICIU}{MICIU/AEI/10.13039/501100011033}{https://doi.org/10.13039/501100011033} and \grantsponsor{ERDF_E}{ERDF, EU}{} under projects \grantnum{PID2022}{PID2022-139293NB-C32} and \grantnum{FPU}{FPU21/01074};
\grantsponsor{UCLM}{Universidad de Castilla-La Mancha}{} and \grantsponsor{ERDF_E}{ERDF, EU}{} under project \grantnum{Grupos}{2022-GRIN-34437}.
\end{acks}

\printbibliography

\end{document}